\documentclass[11pt,a4paper]{article}
\usepackage[hyperref]{naaclhlt2018}
\usepackage{times}
\usepackage{latexsym}
\usepackage{url}
\usepackage{xcolor}
\usepackage{graphicx}
\usepackage{hhline}
\usepackage{amsmath}

\aclfinalcopy

\title{Deep learning evaluation using deep linguistic processing}

\author{
Alexander Kuhnle \\
Department of Computer Science \\
and Technology \\
University of Cambridge \\
{\tt aok25@cam.ac.uk} \\\And
Ann Copestake \\
Department of Computer Science \\
and Technology \\
University of Cambridge \\
{\tt aac10@cam.ac.uk} \\
}

\date{}

\begin{document}

\maketitle

\begin{abstract}
We discuss problems with the standard approaches to evaluation for tasks like visual question answering, and argue that artificial data can be used to address these as a complement to current practice. We demonstrate that with the help of existing `deep' linguistic processing technology we are able to create challenging abstract datasets, which enable us to investigate the language understanding abilities of multimodal deep learning models in detail, as compared to a single performance value on a static and monolithic dataset.
\end{abstract}

\section{Introduction \& related work}

In recent years, deep neural networks (DNNs) have established a new level of performance for many tasks in natural language processing (NLP), speech, computer vision and artificial intelligence. Simultaneously, we observe a move towards simulated environments and artificial data, particularly in reinforcement learning \cite{Bellemare2013,Brockman2016}. As outlined by \newcite{Kiela2016}, simulated data is appealing for various reasons. Most importantly, it acts as a prototypical problem presentation, abstracted from its noisy and intertwined real-world appearance.

However, with a few notable exceptions \cite{Scheffler2001,Byron2007}, artificial data is relatively little used in NLP. Only recently people started arguing for the use of simulated data, like the long-term research proposal of \newcite{Mikolov2015} on learning to understand language from scratch in a virtual environment, and introduced benchmark datasets, like the bAbI tasks \cite{Weston2015} or the VQA datasets discussed below. Here we focus on the problem of \emph{visually grounded language understanding} in the context of visual question answering (VQA). In principle, this task is particularly interesting from a semantic perspective, since it combines general language understanding, reference resolution and grounded language reasoning in a simple and clear task. However, recent work \cite{Goyal2017,Agrawal2016} has suggested that the popular VQA Dataset \cite{Antol2015} is inadequate, due to various issues which allow a system to achieve competitive performance without truly learning these abilities.

To address this, modifications to the existing VQA Dataset and several artificial VQA datasets have been released. The former include C-VQA \cite{Agrawal2017}, a new composition-focused split, and VQA 2.0 \cite{Goyal2017}, an extension based on minimal image pairs. Similar approaches have been proposed in the context of image captioning \cite{Shekhar2017,Hodosh2016}, which relate to our proposal in that they modify language in a principled way. However, despite `mild artificiality', some issues with real-world data like the VQA Dataset remain.

On the other hand, examples of new artificial datasets include the SHAPES dataset \cite{Andreas2016a}, the CLEVR dataset \cite{Johnson2017a}, the NLVR dataset \cite{Suhr2017}, and the ShapeWorld framework \cite{Kuhnle2017}, which is our implementation of the proposal presented here. They all consist of images showing abstract scenes with colored objects and, except for NLVR, use artificially produced language. Language generation for SHAPES and CLEVR is template-based and dataset-specific, while ShapeWorld leverages an existing broad-coverage semantic grammar formalism.

These datasets are introduced with the motivation to provide a clear and challenging evaluation for VQA systems. \newcite{Johnson2017a} and \newcite{Kuhnle2017} investigated popular VQA systems on their datasets, and demonstrate how artificial data provides us with detailed insights previously not possible. Despite its simplicity, they uncover fundamental shortcomings of current VQA models. Since then, CLEVR has been of great importance for the development of new VQA models based on dynamically assembled modules \cite{Hu2017,Johnson2017b}, a dedicated relational module \cite{Santoro2017}, or a general modulation technique \cite{Perez2018}, all of which achieve close-to-perfect accuracy on CLEVR.

The advantage of artificial data in this paper is not seen in its capacity to improve existing models by augmenting training data, although this would be conceivable. Instead we are interested in its capacity to provide data for targeted investigations of specific model capabilities. We argue that it constitutes a \emph{necessary}, though not in itself \emph{sufficient} benchmark for genuine language understanding abilities. The aforementioned models exhibit clearly superior understanding of the types of questions CLEVR contains. This paper proposes a principled way of continuing the incremental progress in multimodal language understanding initiated by CLEVR and its template-based generation approach, based on deep linguistic processing tools. Our initial experiments show that we can provide data that is challenging for state-of-the-art models, like the quantification examples presented in section \ref{section:example}. Note that while success on such narrower datasets may not directly translate to improved performance on broader datasets like the VQA Dataset, the underlying mechanisms are important for progress in the longer run.

Our aims in this paper are threefold. First, we provide a brief but systematic review of the problems surrounding current standard evaluation practices in deep learning. Secondly, we use this to motivate the potential of artificial data from simulated microworlds to investigate DNNs for visually grounded language understanding. Thirdly, we present an evaluation methodology based on linguistic processing resources, and show why compositional semantic representations from a symbolic grammar are particularly suitable for the production of artificial datasets.

\section{Problems of real-world datasets}

In the following, we review a variety of issues related to the practice of evaluating DNNs on popular real-world datasets for tasks like VQA. We emphasize that our arguments are mainly based on large-scale and broad-coverage datasets obtained in a relatively unconstrained way. Some of the points do not (fully) apply to more specific and carefully obtained data, like the NLVR dataset.

\subsection{Issues with crowdsourced real-world data}

The fact that DNNs require immense amounts of data for successful training led to the practice of adopting online data, such as the Flickr photo sharing platform, and leveraging crowdsourcing, usually via Amazon Mechanical Turk (AMT). For instance, the image captioning dataset MS COCO \cite{Lin2014} contains more than 300,000 images annotated with more than 2 million human-written captions, while the popular VQA Dataset \cite{Antol2015} is based on MS COCO.

Data obtained this way tends to be comparatively simple in terms of syntax and compositional semantics, despite exhibiting a high degree of lexical complexity due to its real-world breadth. Moreover, `re-purposed' photos do not -- and were never intended to -- reflect the visual complexity of every-day scenarios \cite{Pinto2008}. Humans given the task of captioning such images will mostly produce descriptions which are syntactically simple. The way that workers on crowdsourcing platforms are paid gives them an incentive to come up with captions quickly, and hence further increases the tendency to simplicity. Note also that, while this is a form of real-world data, it has very little relationship to the way that a human language learner perceives the world, from the fact that image/question pairs are presented in no meaningful order to the impossibility of any kind of interaction with a particular scene.

Natural language follows Zipf's law for many aspects (sentence length, syntactic complexity, word usage, etc), and consequently has an inbuilt simplicity bias when considered in terms of probability mass. The contents of image datasets based on photos also have a Zipfian distribution, but with biases which relate to what people choose to photograph rather than to what they see. Animal images in the VQA Dataset are predominantly cats and dogs, sport images mainly baseball and tennis (see \newcite{Antol2015} for more statistics). Considering all these biases both in language and vision, the common evaluation measure -- simple accuracy of questions answered correctly -- is not a good reflection of a system's general ability to understand visually grounded language.

\subsection{The Clever Hans effect}

Crowdsourced visual questions have other unexpected properties. \newcite{Goyal2017} and \newcite{Mahendru2017} note how questions rarely talk about objects that are not present in the image, hence an existential question like \textit{``Do you see a...?''} is mostly true. \newcite{Agrawal2016} also give the example of questions like \textit{``What covers the ground?''}, which can confidently be answered with \textit{``snow''} because of biases in common real-world scenes, or, more precisely, biases in the photographs of real-world scenes. Such biases help to explain why some \emph{text-only} systems turn out to perform well on \emph{visual} question answering when evaluated on the VQA Dataset.

\newcite{Sturm2014} compared such unexpected cues when evaluation machine learning systems to the story of `Clever Hans', a horse exhibited in the early 20th century which was claimed to understand German and have extensive arithmetical and reasoning abilities. Hans was eventually found to be picking up on very subtle cues which were given completely unconsciously by his owner and which were not noticed by ordinary observers. Some of the recent findings for DNNs, particularly in NLP, suggest similarly problematic conclusions, like the surprisingly strong performance of a bag-of-words model for sequential information \cite{Adi2017} or of text-only systems for multimodal information \cite{Jabri2016}.

A more fundamental form of this effect is illustrated by recent investigations in image recognition. \newcite{Szegedy2014} and \newcite{Nguyen2015} have shown surprisingly odd system behavior when confronted with either only minimally modified images or almost random noise. This behavior seems due to the specific interplay of a few parameters which dominate the model's decision, and have led to an entire research subfield on adversarial instances in vision. Such investigations are not yet as prominent in the NLP community, although see, e.g., \newcite{Jia2017}, \newcite{Sproat2016} and \newcite{Arthur2016}.

The ability to work with raw input data and to pick up correlations/biases, which humans cannot always manifest in explicit symbolic rules, is precisely the strength of DNNs as feature extractors. But given the often millions of parameters and large number of unstructured input values, it is difficult to avoid unexpected hidden cues. Real-world data with its enormous `sample space', which is necessarily only sparsely reflected, is hence particularly prone to this effect.

The immediate problem is that a system trained this way may not generalize appropriately to other situations. The longer-term problem is that, while we do not expect that DNNs will simulate human capabilities in a fine-grained way, there has to be some degree of comparability if they are ever to be capable of justifying or explaining their behavior. The `Clever Hans effect' thus refers to situations where we wrongly and prematurely attribute such human-like reasoning mechanisms to trained models, when more careful and systematic investigations would have revealed our misjudgement.

\begin{figure*}
\centering

\textit{A pentagon is above a green ellipse, and no blue shape is an ellipse.}
\vspace{0.1cm}

\includegraphics[width=\linewidth]{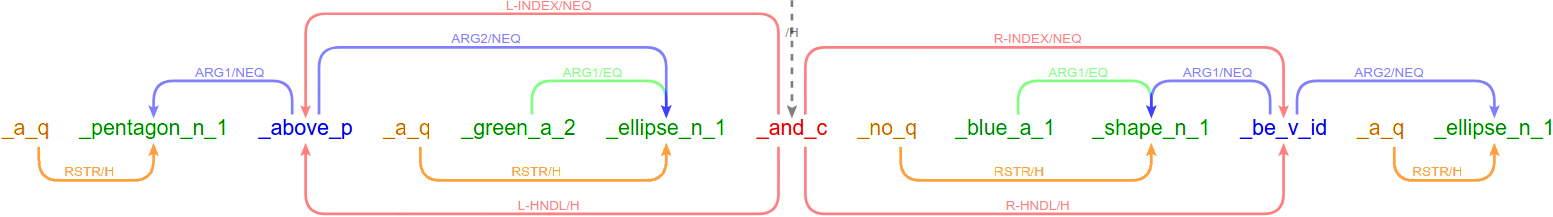}

\fcolorbox{black!50!white}{white}{\begin{minipage}[c][0.18cm][c]{0.03\linewidth}\centering
$\scriptscriptstyle \exists a$
\end{minipage}}%
\fcolorbox{black!50!white}{white}{\begin{minipage}[c][0.18cm][c]{0.097\linewidth}\centering
$\scriptscriptstyle a.\text{shape} = \text{pg}$
\end{minipage}}%
\fcolorbox{black!50!white}{white}{\begin{minipage}[c][0.18cm][c]{0.07\linewidth}\centering
$\scriptscriptstyle a.\text{y} > b.\text{y}$
\end{minipage}}%
\fcolorbox{black!50!white}{white}{\begin{minipage}[c][0.18cm][c]{0.033\linewidth}\centering
$\scriptscriptstyle \exists b$
\end{minipage}}%
\fcolorbox{black!50!white}{white}{\begin{minipage}[c][0.18cm][c]{0.078\linewidth}\centering
$\scriptscriptstyle b.\text{color} = \text{gr}$
\end{minipage}}%
\fcolorbox{black!50!white}{white}{\begin{minipage}[c][0.18cm][c]{0.08\linewidth}\centering
$\scriptscriptstyle b.\text{shape} = \text{el}$
\end{minipage}}%
\fcolorbox{black!50!white}{white}{\begin{minipage}[c][0.18cm][c]{0.05\linewidth}\centering
$\scriptscriptstyle \wedge$
\end{minipage}}%
\fcolorbox{black!50!white}{white}{\begin{minipage}[c][0.18cm][c]{0.047\linewidth}\centering
$\scriptscriptstyle \neg \exists c$
\end{minipage}}%
\fcolorbox{black!50!white}{white}{\begin{minipage}[c][0.18cm][c]{0.075\linewidth}\centering
$\scriptscriptstyle c.\text{color} = \text{bl}$
\end{minipage}}%
\fcolorbox{black!50!white}{white}{\begin{minipage}[c][0.18cm][c]{0.081\linewidth}\centering
$\scriptscriptstyle \text{true}$
\end{minipage}}%
\fcolorbox{black!50!white}{white}{\begin{minipage}[c][0.18cm][c]{0.06\linewidth}\centering
$\scriptscriptstyle c = d$
\end{minipage}}%
\fcolorbox{black!50!white}{white}{\begin{minipage}[c][0.18cm][c]{0.034\linewidth}\centering
$\scriptscriptstyle \exists d$
\end{minipage}}%
\fcolorbox{black!50!white}{white}{\begin{minipage}[c][0.18cm][c]{0.07\linewidth}\centering
$\scriptscriptstyle d.\text{shape} = \text{el}$
\end{minipage}}%
\\[-0.06cm]
\fcolorbox{black!50!white}{white}{\begin{minipage}[c][0.18cm][c]{0.142\linewidth}\centering
$\scriptscriptstyle \exists a \colon\, a.\text{shape} = \text{pg}$
\end{minipage}}%
\fcolorbox{black!50!white}{white}{\begin{minipage}[c][0.18cm][c]{0.07\linewidth}\centering
$\scriptscriptstyle a.\text{y} > b.\text{y}$
\end{minipage}}%
\fcolorbox{black!50!white}{white}{\begin{minipage}[c][0.18cm][c]{0.221\linewidth}\centering
$\scriptscriptstyle \exists b \colon\, b.\text{color} = \text{gr} \:\wedge\: b.\text{shape} = \text{el}$
\end{minipage}}%
\fcolorbox{black!50!white}{white}{\begin{minipage}[c][0.18cm][c]{0.05\linewidth}\centering
$\scriptscriptstyle \wedge$
\end{minipage}}%
\fcolorbox{black!50!white}{white}{\begin{minipage}[c][0.18cm][c]{0.233\linewidth}\centering
$\scriptscriptstyle \neg \exists c \colon\, c.\text{color} = \text{bl}$
\end{minipage}}%
\fcolorbox{black!50!white}{white}{\begin{minipage}[c][0.18cm][c]{0.06\linewidth}\centering
$\scriptscriptstyle c = d$
\end{minipage}}%
\fcolorbox{black!50!white}{white}{\begin{minipage}[c][0.18cm][c]{0.119\linewidth}\centering
$\scriptscriptstyle \exists d \colon\, d.\text{shape} = \text{el}$
\end{minipage}}%
\\[-0.06cm]
\fcolorbox{black!50!white}{white}{\begin{minipage}[c][0.18cm][c]{0.463\linewidth}\centering
$\scriptscriptstyle \exists a \colon\, a.\text{shape} = \text{pg} \:\wedge\: \left[ \exists b \colon\, b.\text{color} = \text{gr} \:\wedge\: b.\text{shape} = \text{el} \:\wedge\: a.\text{y} > b.\text{y} \right]$
\end{minipage}}%
\fcolorbox{black!50!white}{white}{\begin{minipage}[c][0.18cm][c]{0.05\linewidth}\centering
$\scriptscriptstyle \wedge$
\end{minipage}}%
\fcolorbox{black!50!white}{white}{\begin{minipage}[c][0.18cm][c]{0.442\linewidth}\centering
$\scriptscriptstyle \neg \exists c \colon\, c.\text{color} = \text{bl} \:\wedge\: \left[ \exists d \colon\, d.\text{shape} = \text{el} \:\wedge\: c = d \right]$
\end{minipage}}%
\\[-0.06cm]
\fcolorbox{black!50!white}{white}{\begin{minipage}[c][0.18cm][c]{0.985\linewidth}\centering
$\scriptscriptstyle \left( \exists a \colon\, a.\text{shape} = \text{pg} \:\wedge\: \left[ \exists b \colon\, b.\text{color} = \text{gr} \:\wedge\: b.\text{shape} = \text{el} \:\wedge\: a.\text{y} > b.\text{y} \right] \right) \;\wedge\; \left( \neg \exists c \colon\, c.\text{color} = \text{bl} \:\wedge\: \left[ \exists d \colon\, d.\text{shape} = \text{el} \:\wedge\: c = d \right] \right)$
\end{minipage}}
\caption{A caption with DMRS graph and semantic interpretation, illustrating how compositionality enables us to generate combinatorially large amounts of non-trivial captions and infer their semantics from atomic elements.}
\label{figure:dmrs}
\end{figure*}

\subsection{Guiding principles for DNN evaluation}
Compositionality is a fundamental aspect of language, and consequently a necessary prerequisite for any claim about \emph{`understanding'} natural language. Besides being required for proper generalization to novel utterances, it constitutes a far more efficient way of learning in the form of a structural prior, it leads to more interpretable inference results by forcing more systematic processing, and it results in more robust behavior, promising to reduce vulnerability to adversarial examples. However, \newcite{Lake2017} recently gave reason to doubt the compositional capabilities of recurrent DNNs, which are at the heart of virtually all state-of-the-art NLP models. We conclude from this that a different kind of test data than existing benchmarks is required for more conclusive evaluations, and propose three simple principled ways to reduce the risk of encountering such problems:
\begin{itemize}
\setlength{\topsep}{0cm}
\setlength{\parsep}{0cm}
\setlength{\itemsep}{0cm}
\item Avoid solely evaluating a system on a single and supposedly representative set, but design test instances with the aim of specifically investigating and confirming the system's intended improvement over other models.
\item Instead of keeping training and test data distributions similar, focus on the true compositional generalization abilities required by dissimilar distributions. A more asymmetric dataset represents a harder, but hence potentially more interesting task.
\item Do at least some experiments with clean data, which reduces the likelihood of hidden biases or correlations compared to more `realistic' and complex data. For instance, the relationship between image and text should be explicitly controlled in multimodal data.
\end{itemize}

\section{Automatic generation of language data}

In the following, we describe our approach for automatic generation of artificial VQA data using existing deep linguistic processing technology, based on our implementation in the ShapeWorld framework \cite{Kuhnle2017}\footnote{\scriptsize\url{https://github.com/AlexKuhnle/ShapeWorld}}. We argue that a compositional semantic approach using a bidirectional grammar gives us precisely the sort of data as outlined by the above principles. We propose this approach as a complementary evaluation step, since it is not intended to replace real-world evaluation, but instead aims to cover aspects which existing datasets cannot provide.

\subsection{Abstract microworlds}

The generation process we use is based on randomly sampled abstract world models, i.e.\ values which specify a microworld, entities and all their attributes. In the case of our framework these include the number of entities, their shape and color, position, rotation, shade, etc. Such a world model can be visualized straightforwardly.

In this context, datasets are generators which can create an unlimited amount of data instances, hence making multiple iterations over a fixed set of training instances obsolete. Importantly, different datasets constrain the general sampling process in different ways by, for instance, restricting the number of objects, the attribute values available, the positioning of entities, and more. This addresses the point of specifying different data distributions for training and testing. Moreover, it makes it possible to partition evaluation data as desired, which facilitates the detailed investigation of system behavior for specific instances and hence the discovery of systematic shortcomings.

\begin{figure*}
\newcommand{\correct}[1]{\textcolor{green!60!black}{#1}}
\newcommand{\incorrect}[1]{\textcolor{red!80!black}{#1}}
\centering
\begin{minipage}{0.175\linewidth}
\includegraphics[width=\linewidth]{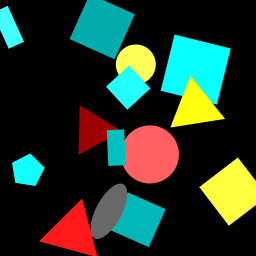}
\end{minipage}%
\begin{minipage}{0.5\linewidth}
\footnotesize
\begin{itemize}
\setlength{\parsep}{0cm}
\setlength{\itemsep}{0cm}
\item \correct{Less than one triangle is cyan.}
\item \correct{At least half the triangles are red.}
\item \incorrect{More than a third of the shapes are cyan squares.}
\item \incorrect{Exactly all the five squares are red.}
\item \correct{More than one of the seven cyan shapes is a cyan square.}
\item \incorrect{Twice as many red shapes as yellow shapes are circles.}
\end{itemize}
\end{minipage}
\caption{An image with several example captions focusing on quantification. The task is image caption agreement, that is, to decide whether the caption agrees with the image (\correct{green}) or not (\incorrect{red}), similar to yes/no questions in VQA.}
\label{figure:example}
\end{figure*}

\subsection{Syntactically rich language generation}

Of the recent abstract datasets mentioned in the introduction, \newcite{Suhr2017} use human-written captions, the SHAPES dataset \cite{Andreas2016a} a minimalist grammar, and the CLEVR dataset \cite{Johnson2017a} a more complex one based on functional building blocks, both template-based and specifically designed for their data. For our approach we leverage technology made available by the DELPH-IN (Deep Linguistic Processing with HPSG) consortium. More specifically, we make use of the broad-coverage, bidirectional\footnote{Bidirectional grammars can be used for generation as well as parsing, of which the latter might be useful here, for instance, in investigating ambiguity effects.},
high-precision English Resource Grammar \cite{Flickinger2000}, which builds on the compositional semantic framework of Minimal Recursion Semantics \cite{Copestake2005}. For our system we use one of its variant, Dependency MRS (DMRS, \newcite{Copestake2009}, \newcite{Copestake2016}), and generate natural language sentences from abstract DMRS graphs using Packard's parser-generator ACE\footnote{\scriptsize\url{http://sweaglesw.org/linguistics/ace/}}.

We have found that DMRS graphs can easily be enriched with appropriate semantics to be evaluated on a given world model. This means that the internals of the language system are essentially using a form of model-theoretic semantics. However, the external presentation of our task is still \textit{`natural'}, i.e.\ only consists of image and language. Compositional representations like DMRS further enable us to produce an infinite number of captions of arbitrary syntactic complexity.

Figure \ref{figure:dmrs} shows an example of a non-trivial caption with corresponding DMRS graph and logical representation over a world model. Both the abstractness and compositionality of the semantic representation are essential to allow us to scale beyond toy examples. The abstract scenario puts an emphasis on experiments with closed-class vocabulary and syntax, as compared to open-class dominated real-world datasets. However, the same approach can be extended to more complex domains, like the clip-arts of \newcite{Zitnick2016}.

In the future, we plan to implement two interesting extensions for our framework: First, paraphrase rules can be expressed on grammar-level and integrated into the generation process as post-processing step for increased linguistic variety. Second, (D)MRS-based grammars for other languages, such as the JACY grammar for Japanese \cite{Siegel2016}, can be used simply by translating the internal mapping of atomic DMRS components to corresponding semantic elements.

\subsection{Quantification example}\label{section:example}

Figure \ref{figure:example} presents an example image accompanied by a variety of correct and incorrect captions focusing on quantification. We produce both count-based (\textit{``three''}) and fraction-based (generalized) quantifiers (\textit{``half''}) in various modifications (\textit{``less than three''}), optionally with additional number restriction (\textit{``at least three of the five''}) or comparative (\textit{``half as many as''}).

We decide to focus on quantifiers here because, on the one hand, they can exhibit a high degree of structural complexity, which can only be resolved by using visual information. On the other hand, categories like `number' in VQA 2.0 or `count' and `compare integer' in CLEVR imply that count-based quantification is specifically covered by these datasets. As the various captions in figure \ref{figure:example} illustrate, this is not fully the case. Note that we so far do not consider scope ambiguity of nested quantifiers, although our approach can be extended accordingly, since the (D)MRS formalism supports scope underspecification, which is one of the reasons for choosing DMRS.

\section{Conclusion: Why use artificial data?}


\paragraph{Challenging test data.} The interplay of abstract world model and semantic language representation enables us to generate captions requiring non-trivial multimodal reasoning. In fact, the resulting captions can be more complex than the sort of captions we could plausibly obtain from humans, and do not suffer from a Zipfian tendency to simplicity on average (unless configured accordingly).

\paragraph{Avoid Clever Hans effect.} The simple, abstract domain and the controlled generation process based on randomly sampling microworlds makes such data comparatively unbiased and greatly reduces the possibility of hidden complex correlations. We can be confident that we cover the data space both relatively uniformly and more exhaustively than this is the case in real-world datasets.

\paragraph{Flexibility \& reusability.} Real-world and/or human-created data essentially has to be obtained again for every change/update, like for VQA v2.0 \cite{Goyal2017}.
In contrast to that, modularity and detailed configurability make our approach easily reusable for a wide range of potentially unforeseen changes in evaluation focus.

\paragraph{Rich evaluation.} Ultimately, our goal in providing datasets is to enable detailed evaluations (of DNNs). By creating atomic test datasets specifically evaluating instance types individually (e.g., counting, spatial relations, or even more fine-grained), we can unit-test a DNN for specific subtasks. We believe that such a modular approach is a better way to establish trust in the understanding abilities of DNNs than a monolithic dataset and a single accuracy number to assess performance.

\section*{Acknowledgments}

We thank the anonymous reviewers for their constructive feedback. AK is grateful for being supported by a Qualcomm Research Studentship and an EPSRC Doctoral Training Studentship.

\bibliography{bibliography}
\bibliographystyle{acl_natbib}

\end{document}